\documentclass[manuscript]{acmart}
\usepackage{algorithm} 
\usepackage{algpseudocode}

\AtBeginDocument{%
  \providecommand\BibTeX{{%
    \normalfont B\kern-0.5em{\scshape i\kern-0.25em b}\kern-0.8em\TeX}}}
\newtheorem{prop}{Property}

\setcopyright{acmcopyright}
\copyrightyear{2018}
\acmYear{2018}
\acmDOI{XXXXXXX.XXXXXXX}

\acmConference[Conference acronym 'XX]{Make sure to enter the correct
  conference title from your rights confirmation emai}{June 03--05,
  2023}{Woodstock, NY}
\acmPrice{15.00}
\acmISBN{978-1-4503-XXXX-X/18/06}

\begin{document}

\title{Transfer: Cross Modality Knowledge Transfer using Adversarial Networks - A Study on Gesture Recognition}


\author{Payal Kamboj*}
\affiliation{%
  \institution{Arizona State University}
 \city{Tempe}
  \state{Arizona}
  \country{USA}}
\email{pkamboj@asu.edu}

\author{Ayan banerjee*}
\email{abanerj3@asu.edu}
\affiliation{%
  \institution{Arizona State University}
  \city{Tempe}
  \state{Arizona}
  \country{USA}
  \postcode{85281}
}
 \email{abanerj3@asu.edu}

\author{Sandeep K.S. Gupta}
\affiliation{%
\institution{Arizona State University}
  \city{Tempe}
    \state{Arizona}
  \country{USA}}
  \email{Sandeep.Gupta@asu.edu}


\begin{abstract}
  Knowledge transfer across sensing technology is a novel concept that has been recently explored in many application domains, including gesture-based human computer interaction. The main aim is to gather semantic or data driven information from a source technology to classify / recognize instances of unseen classes in the target technology. The primary challenge is the significant difference in dimensionality and distribution of feature sets between the source and the target technologies. In this paper, we propose \textit{TRANSFER}, a generic framework for knowledge transfer between a source and a target technology. \textit{TRANSFER} uses a language-based representation of a  hand gesture, which captures a temporal combination of concepts such as handshape, location, and movement that are semantically related to the meaning of a word. By utilizing a pre-specified syntactic structure and tokenizer, \textit{TRANSFER} segments a hand gesture into tokens and identifies individual components using a token recognizer. The tokenizer in this language-based recognition system abstracts the low-level technology-specific characteristics to the machine interface, enabling the design of a discriminator that learns technology-invariant features essential for recognition of gestures in both source and target technologies. We demonstrate  the usage of \textit{TRANSFER} for three different scenarios: a) transferring knowledge across technology by learning gesture models from video and recognizing gestures using WiFi, b) transferring knowledge from video to accelerometer, and d) transferring knowledge from accelerometer to WiFi signals.
\end{abstract}







\maketitle
* These authors contributed equally to this work.
\section{Introduction}
\label{sec:intro}
Gesture recognition is a crucial aspect of mobile Human-Computer Interaction (HCI) applications such as sign language recognition~\cite{gope2012hand} ~\cite{1}  ~\cite{2}~\cite{3} , smart home gesture interfaces~\cite{kuhnel2011m}, gesture-based control of devices~\cite{lu2014hand}, and monitoring of hygiene practices~\cite{moore2021impact}. Device-free gesture recognition is essential for supporting ubiquitous HCI\footnote{The definition of device free gesture recognition almost exclusively means no device on the human body. To the best of our knowledge all works which claim to be device free require installation of specialized equipment in the environment. Even the works claiming the usage of commodity over the shelf (COTS) devices require installation of WiFi access points at strategic positions. The only exception is video based recognition systems that can operate using COTS devices such as user's smartphone.} using WiFi\cite{Ma,Zou,Wigrunt,Regani,Zhou,wicatch,WiDraw,wigest,WiHF,ma2018signfi}, acoustic signals\cite{soundwrite,HCI,Push,Wang}, RFID\cite{GRfid,Headsee} and video\cite{smith2021improved,Paudyal16Sceptre}. However, in many cases, proposed technologies are tested on a limited set of gestures\cite{Sameena2023}. 
On average, the number of gestures used is only nine, and the number of users is usually fewer than fifteen. Notable exceptions to this trend include SignFi\cite{ma2018signfi}, a WiFi based gesture recognition mechanism that tests on 276 American Sign Language (ASL) gestures, albeit from only one user, and Paudyal et al.~\cite{paudyal2020towards}, a video-based system that tests on 25 gestures on 100 users. 
\begin{figure*}[h]
	\centering
	\includegraphics[width=\linewidth]{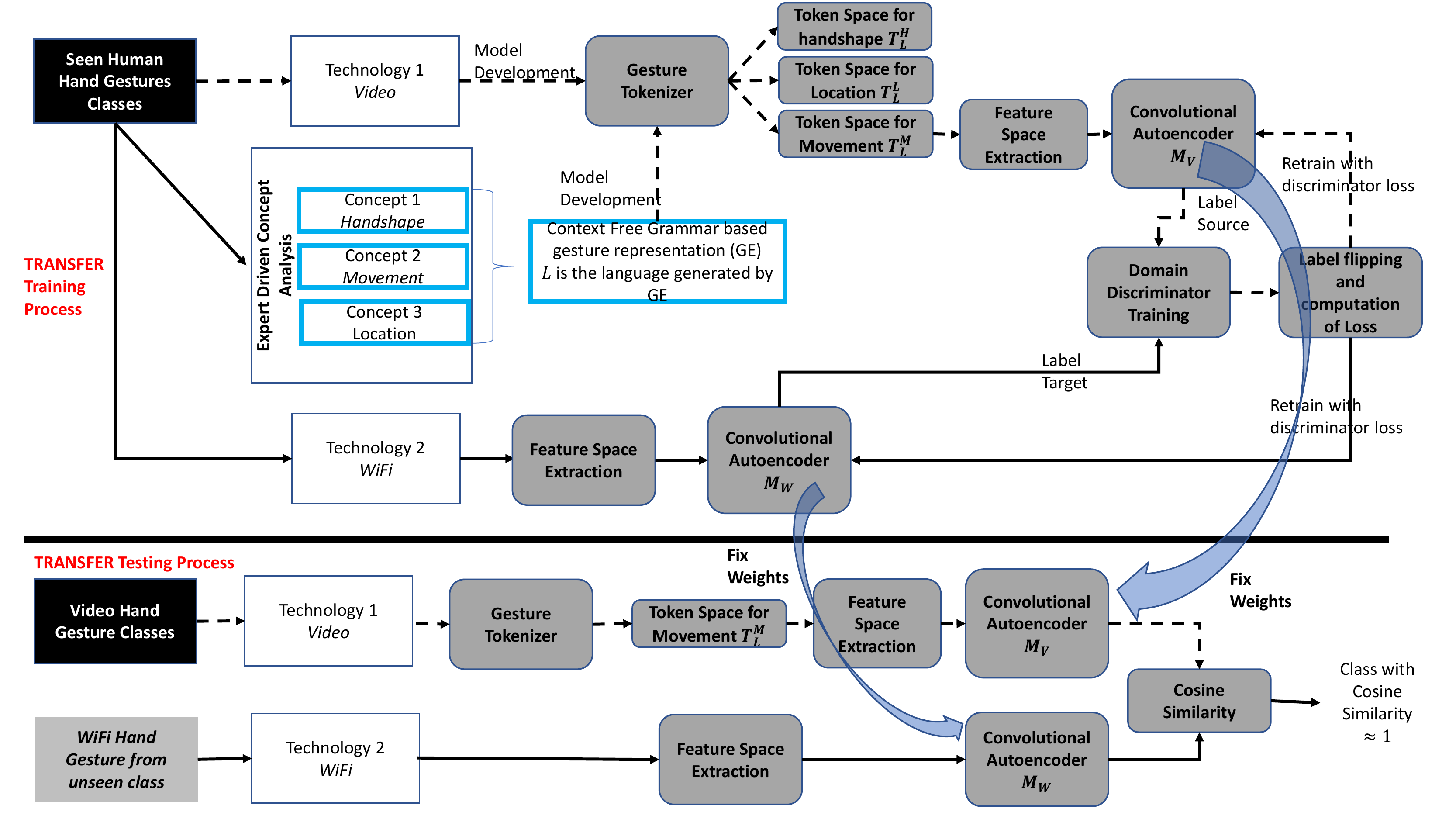}
	\caption{\textit{TRANSFER} framework internals and interfaces explained using source domain as Video and Target domain as WiFi. Similar approach is explored in the paper with source as Video and target as accelerometer. }
	\label{fig:MUDRAFig}
	\Description{}
\end{figure*}
The recognition of the gestures in ubiquitous HCI applications typically involves a supervised classification step~\cite{smith2021improved},  which requires data collection using a specific sensing technology to train the classification model. This data collection step can be time-consuming, resource-intensive, sensing technology-dependent, and may limit the use of high-performance machine learning techniques for recognition. In this paper, we propose \textit{TRANSFER}, which aims to enable gesture recognition beyond sensing barriers.  Specifically, we use data collected using one sensing technology to extract representational knowledge in a latent parameter space and apply it to recognize gestures using a different sensing technology. We refer to this as \textit{Technology Adaptation} (TA), a form of cross-modal learning that enables the transfer of knowledge across different sensing technologies with inconsistent distributions and representations. 

\subsection{Need for \textit{TRANSFER}} 
Human gesture recognition using video sensing is a well established area with several publicly available large-scale datasets such as SignSavvy~\cite{Savvy}, ASLTEXT~\cite{bilge2019zero} for American Sign Language (ASL)~\cite{b38},  military hand gestures~\cite{lin2009recognizing} and air marshal hand gestures for human action recognition. Video-based gesture recognition has been shown to have good \textit{cross-domain} accuracy, meaning it is robust across environments, devices, users, locations, and orientations~\cite{b39}~\cite{b40}~\cite{b41}~\cite{b42}. However, in practice, the recognition of gestures or actions using video has a major privacy violation risk. To overcome such issues, newer sensing technologies have been recently explored such as WiFi, RFID, mmWave, sEMG, and accelerometer (Table \ref{tbl:RelWork}). Some are device free such as WiFi and mmWave, and require infrastructure overhead, while others require a device to be worn by the user such as sEMG and accelerometer. An important drawback of these non-video based techniques is that they have poor cross-domain performance (even changing the user drastically reduces accuracy)~\cite{Wilson}, and data collection is expensive since typically they require precise placement of sensing devices in each new environment. Utilizing knowledge is key to addressing this problem effectively~\cite{Banerjee2023}~\cite{Kamboj2023}. One of the main achievements of \textit{TRANSFER} is its ability to utilize the knowledge embedded in the vast latent space of video data, which has been trained by a plethora of databases. This enables good cross-technology gesture recognition performance with non-video sensing techniques, without the need for additional data collection.
\begin{table}[t]
	\sffamily
	\centering
	\footnotesize
	\caption{Comparison of non-Video Device Free Gesture Recognition works w.r.t number of gestures, number of users, and cross domain applicability. NS: Not Specified }
	\scriptsize
	\begin{tabular}{p{0.7 in}|p{0.4 in}|p{0.7 in}|p{0.3 in}|p{0.3 in}|p{0.3 in}} \toprule
\textbf{Work}		&{\bf Technology} & {\bf Gestures} &{\bf Users} & {\bf Cross Technology } & {\bf Accuracy}\\ \midrule
		Wang et al \cite{Wang} & Acoustic & 26 alphabets, 11 words & 5 & No & 92.3\% and 91.2\% \\  
		Ma et al \cite{Ma} & WiFi & 26 alphabets & 14 & No & 93.2\%\\
		Zou et al \cite{GRfid} & RFID & 6 hand movements & 4 & No & 92.8\%\\
		Zou et al \cite{Zou} & WiFi & 6 hand movements & 1 & No & 96.5\%\\
		WiGRUNT\cite{Wigrunt} & WiFi & 6 hand movements & 6 & No & 99.7\%\\
		Zhang et al\cite{Zhang} & WiFi and mmWave & 6 hand movements & 16 & No & 94.9\%\\
		Regani et al \cite{Regani} & WiFi & 6 uppercase alphabets & 1 & No & 87\%\\
		Wang et al \cite{Wang2} & mmWave & 8 & 4 & No & 96.3\%\\
		Multi-Touch  \cite{Wang3} & RFID & 26 alphabets & 10 & No & 80\%\\
		Zhou et al \cite{Zhou} & WiFi & 10 digits & 5 & No & 94\% \\
		SoundWrite \cite{soundwrite} & Acoustic & 7 hand movements & 10 & No & 91\%\\
		WiCatch \cite{wicatch} & WiFi & 9 hand movements & 1 & No & 96\% \\
		SMART \cite{Smart} & Ambient Light Sensor & 9 hand movements & 8 & No & 96\% \\
		WiDraw \cite{WiDraw} & WiFi & 26 alphabets, 1000 words & 10 & No & 85\% \\
		Luo et al (HCI) \cite{HCI} & Acoustic & 7 hand movements & 10 & No & 93.2\%\\
		WiGest \cite{wigest} & WiFi & 5 hand movements & 1 & No & 96 \% \\
		Wang et al \cite{Push} & Acoustic & 15 hand movements & 8 & No & 96\%\\
		WiHF \cite{WiHF} & WiFi & 6 hand movements & 6 & No & 97\%\\
		Han et al \cite{b40} & WiFi & 4 hand movements & 1 & No & 94.1 \% \\
		TeraHertz \cite{WangTera} & Terahertz Radar	& 10 hand movements & 1 & No & 95.5\% \\
		HeadSee \cite{Headsee} & RFID & 10 hand movements & 5 & No & 91 \% \\
		PWiG \cite{pwig} & WiFi & 5 hand movements & 1 & No & 89\%\\  
		SignFi \cite{ma2018signfi} & WiFi & 276 ASL & 5 & No & 86\%\\ 
        CrossGR \cite{CrossGRLi2021} &WiFi &15 &10 &No &82.6\%\\
        Wu et al. \cite {WuCM} &Video, Kinematics &15 &NS &Yes &45\%-65\% \\
		\textbf{TRANSFER} & \textbf{WiFi + Video} & \textbf{250 ASL} & \textbf{22 users across datasets} & \textbf{Yes} & \textbf{Table \ref{tbl:Sum}}\\
		\bottomrule
	\end{tabular}
	\label{tbl:RelWork}
\end{table}
\section{Related works}
Table \ref{tbl:RelWork} highlights several attempts at non-video device-free gesture recognition, with the maximum number of gestures used across all works (excluding SignFi) being 15. SignFi stands out as the first work to apply device-free WiFi-based gesture recognition to ASL, utilizing a vast vocabulary of 276 gestures. However, its reported accuracy was 94\% for one user and dropped to 86\% when five users were included. Most of the works in the literature reporting greater than 95\% accuracy are based on a limited number of hand gestures and users. This limitation is mainly due to the time-consuming data collection task that is required beforehand.

Most of the literature work focuses on gesture recognition using a single technology. Wang et al. \cite{Wang} proposes a device-free gesture tracking scheme called LLAP that uses speakers and microphones on mobile devices to track hand/finger movements through acoustic phase measurement. LLAP has a tracking accuracy of 3.5mm and 4.6mm for 1-D and 2-D movements, respectively, and can recognize characters and short words drawn in the air with 92.3\% and 91.2\% accuracy. Ma et al. \cite{Ma} proposes a device-free gesture recognition system based on meta-learning that can recognize new types of gestures or gestures performed in new conditions with an accuracy of over 90\%, using very few new samples. The system utilizes a deep network that learns both discriminative deep features and a transferrable similarity evaluation ability from the training set, enabling it to apply learned knowledge to new testing conditions. Zou et al. \cite{GRfid} proposes GRfid, a device-free gesture recognition system based on phase information output by COTS RFID devices, to overcome the intrusiveness of existing RFID-based gesture recognition systems. GRfid achieved an average recognition accuracy of 96.5\% and 92.8\% in identical-position and diverse-positions scenarios, respectively, and was found to be robust against environmental interference and tag orientations. The system is designed with well-designed functional blocks for high-performance gesture recognition and utilizes low-cost commodity hardware.  Zou et al. \cite{Zou} proposes a WiFi-enabled device-free adaptive gesture recognition scheme, WiADG, that can accurately recognize human gestures under environmental dynamics via adversarial domain adaptation. WiADG achieves 98\% gesture recognition accuracy in the original environment and improves recognition accuracy by 25\% on average when implemented in new environments, without the need for labeled data collection and new classifier generation. The proposed scheme utilizes an OpenWrt-based IoT platform to collect Channel State Information (CSI) measurements from commercial IoT devices and employs a convolutional neural network for accurate source classification. Gu et al.  \cite{Wigrunt} proposes a WiFi-based gesture recognition system called WiGRUNT that uses a dual-attention network to capture domain-independent features of a gesture on WiFi Channel State Information. WiGRUNT outperforms existing state-of-the-art approaches in both in-domain and cross-domain evaluations using the Widar3 dataset. The proposed system can be applied in various applications that require device-free, non-line-of-sight, and privacy-friendly gesture recognition. Similary, there are many works on device-free gesture recognition using different technologies that attempted gesture recognition on limited hand gestures and users using accoustic, WiFi, RFID, mmWave and light sensor signals, however they were confined to either a single domain or if they attempted cross-domain recognition their "domain" term refers to gesture irrelevant factors like locations, users, orientation etc.  Zhang et al. \cite{Zhang}. 

Wu et al. \cite{WuCM} learnt the cross-modal representation of surgical robotic activity from video to kinematics data using self-supervised learning and achieved an accuracy of 45-65\% on unseen gestures in training.

A reason for less number of hand gestures considered in recent work is that non-video data collection technology is not responsive to differences in hand shapes and can only capture arm movements. Most of the works seen in Table \ref{tbl:RelWork} use gestures where handshapes are irrelevant. Works that do recognize English alphabets, consider air drawing and hence do not need hand shapes. 

\section{Problem Definition}
\noindent{\bf Given:} Given a set of gesture classes $\mathcal{G}$, we assume that there are large number of feature samples for every class in the source technology. Hence, we assume a labelled set $(X^S,Y^S)$, where $X^S$ is the feature set in the source technology, while $Y^S$ is the label set for each element in $X^S$. In the target technology, we assume a feature set $X^T$ with two domains: a) a labeled domain $\mathcal{D}_L$, which has labeled feature data $(X^L,Y^L)$ for a subset $\mathcal{G}_L \subset \mathcal{G}$ of gesture classes, and b) an unlabeled domain $\mathcal{D}_u$, which has only feature data $X^u$ without any labels in the target technology. Note that $X^T = X^L \bigcup X^u$.

\noindent{\bf Objective:} To generate a label set $Y^u$ for the unlabeled feature set $X^u$. 
\section{Challenges with Domain Adaptation}
Domain Adaptation (DA) aims to transfer knowledge across different domains that exhibit dissimilar data distributions, but utilize the same technology with similar data dimensions and features. Although neural networks and other machine learning models can effectively generalize when trained on large-scale datasets, their performance can be affected by the covariate shift that arises from discrepancies between the source and target domains  ~\cite{b39,b40,b41,b42,b43}.  It should be noted that DA involves different domains but with the same technology, thus ensuring that the features and dimensions are similar. For instance, electrocardiogram (ECG) data captured from older populations using electrocardiograph machines will likely have a different feature space distribution than that of younger populations. Although the number of features is constant, these differences in feature space distribution pose a major challenge in DA. In classic DA, the source and target domains are assumed to have the same feature dimension $d$. Given $d$ dimensional features ${X} \in \mathcal{R}^d$ and labels ${Y}\in \mathcal{R}$, we consider two subsets: a) labeled source subset, represented by  $(x^L_j,y^L_k)_{k \in \{1 \ldots m_S\}} \in (X^L,Y^L)$, and b) an unlabeled target subset, $x^u_k \in X^u$ such that the joint probability distribution of features and labels of the source subset is not equal to that of the target subset, $P^L(X,Y)\neq P^u(X,Y)$. The goal of domain adaptation is to reduce the distribution shifts between the source and target domains by creating a common feature space. To achieve this, techniques such as minimizing maximum mean discrepancy, minimizing correlation distances, adversarial training, and feature translation are used. By reducing these distribution shifts, it becomes possible to develop accurate predictive models that are generalizable to new domains. 

One key difference between DA and TA is that in the latter case, the source and target technologies may not have the same feature dimensions, i.e., if $X^S \in \mathcal{R}^{d_1}$ and $X^T \in \mathcal{R}^{d_2}$ then $d_1 \neq d_2$.  A potential solution is to utilize representational learning which requires learning a latent representation of the source domain using an autoencoder $\phi^S: {X} \rightarrow Z \rightarrow Y$, and the target domain using a second autoencoder $\phi^T: X \rightarrow Z \rightarrow Y$, where $Z$ is the latent space, such that probability densities for both the sensing technologies satisfy the property $P^S(Z|Y) = P^T(Z|Y)$. Specifically, an autencoder is trained to map the features of each technology to a shared latent space of a chosen dimension. These latent space's feature discrepancies can be minimized by using adversarial training. Adversarial training has recently gained popularity as they seek to minimize the domain discrepancy through an adversarial domain discriminator with the similar idea of generative adversarial networks (GANs). Instead of domain, we will use adversarial training to minimize the technology discrepancies here for the TA.

\section{\textit{TRANSFER} Approach}

The main focus of TA is on improving cross-technology performance for different sensing methods, rather than addressing domain shifts. It has been proposed to improve cross-technology performance of gesture recognition for various sensing methods such as WiFi~\cite{b44,b45} and mmWave~\cite{b46,b47}. In this paper, we introduce the \textit{TRANSFER} framework, which facilitates the transfer of deep knowledge extracted from a high-dimensional sensing technology to a low-dimensional sensing technology, or vice versa, enabling  \textit{seamless technology-independent gesture recognition without the need for rebuilding data-driven models}.

The \textit{TRANSFER} framework adopts a language-driven approach for gesture recognition, as depicted in Figure \ref{fig:MUDRAFig}. It defines gestures as a combination of $p$ concepts that are organized using a grammatical structure $GE$.  Each concept $c^i_{L} \in C_L$ corresponds to a high-level characteristic of a gestural language $L$ and is represented by a dimensionality-reducing \textit{token mapping} $\psi_i: X \rightarrow T^i_L$, where $T^i_L \in \mathcal{R}^{d_{T^i_L}}$ and $d_{T^i_L} < d$. By aggregating all concepts, the framework focuses only on the dimensions that are critical for accurate gesture recognition, and discards the remaining dimensions that are not relevant for the recognition process. The grammar $GE$ and the set of concepts $C_L$ are obtained from expert knowledge of the gestural language. The \textit{token mappings} are utilized to facilitate cross-technology gesture recognition using TA through representational learning techniques.

\textit{TRANSFER} relies on the following property for TA:

\begin{prop} \label{prop:P1}
Given a labeled data from source technology with feature domain $(X^S,Y^S)$ and a target technology with unlabeled feature domain $X^T$, 
$\exists \psi_i: X^S \rightarrow T^i_L$ such that two autoencoders $\phi^S : T^i_L \rightarrow Z \rightarrow Y$ and $\phi^T: X^T \rightarrow Z \rightarrow Y$ can be learned from the source and target technology into a common latent space $Z$ with $P^S(Z|Y) = P^T(Z|Y)$ using adversarial training. 
\end{prop}

Based on Property \ref{prop:P1}, DA-inspired adversarial training techniques~\cite{ketyko2019domain,zou2018robust,lee2020improving} are applied to TA by relearning $\phi^S$ and $\phi^T$ through a discriminator's loss function, resulting in a latent space representation $Z$ where $P^S(Z) \approx P^T(Z)$. This approach is then used to recognize gestures from the target sensing domain.

\subsection{Advantages of \textit{TRANSFER}}

There are several significant advantages of utilizing a language-driven approach for gesture recognition in \textit{TRANSFER}, including:

{\bfseries Technology-independent recognition:} The primary characteristic of \textit{TRANSFER} is that data-driven models can be constructed using data collected from one technology, such as video, and subsequently used to recognize gestures using data collected from another technology, such as WiFi or RFID or acoustic signals. This paper demonstrates knowledge transfer between video and WiFi, which implies that the gesture recognition is device-agnostic, meaning that there is no need for a device to be attached to the body.

{\bfseries User independent recognition:}
One of the advantages of \textit{TRANSFER} is its ability to perform user-independent recognition. By decomposing a gesture into concepts, specialized machine learning techniques can be applied to each concept. For instance, handshape recognition requires image classifiers and models such as the Inception model~\cite{murthy2010hand} enable user-independent classification of handshape. Location classification can be achieved using CNN-based techniques like Posenet~\cite{papandreou2017towards}, which provide (X,Y) coordinates of different keypoints on the body. Movement of the hands can be converted to time series data by extracting Posenet keypoints frame by frame from a video. Normalizing the movement across individuals using a fixed point on the body as a reference enables user-independent recognition of movement in \textit{TRANSFER}. We demonstrate this by learning ASL gesture models from a subset of the ASLText dataset, but recognizing gestures on a non-intersecting subset. 


{\bfseries Anytime anywhere training data collection:} Another advantage of \textit{TRANSFER} is its capability for anytime anywhere training data collection. As it allows technology independent recognition, the training data can be gathered using any sensor, offering flexibility in the data collection process. This expands the potential datasets that can be used for training the \textit{TRANSFER} model, potentially resulting in a more comprehensive and accurate language-based gesture communication translation engine. While this is not the primary focus of the paper, it is a significant potential benefit worth noting.

\begin{figure*}
	\centering
	\includegraphics[width=0.8\linewidth]{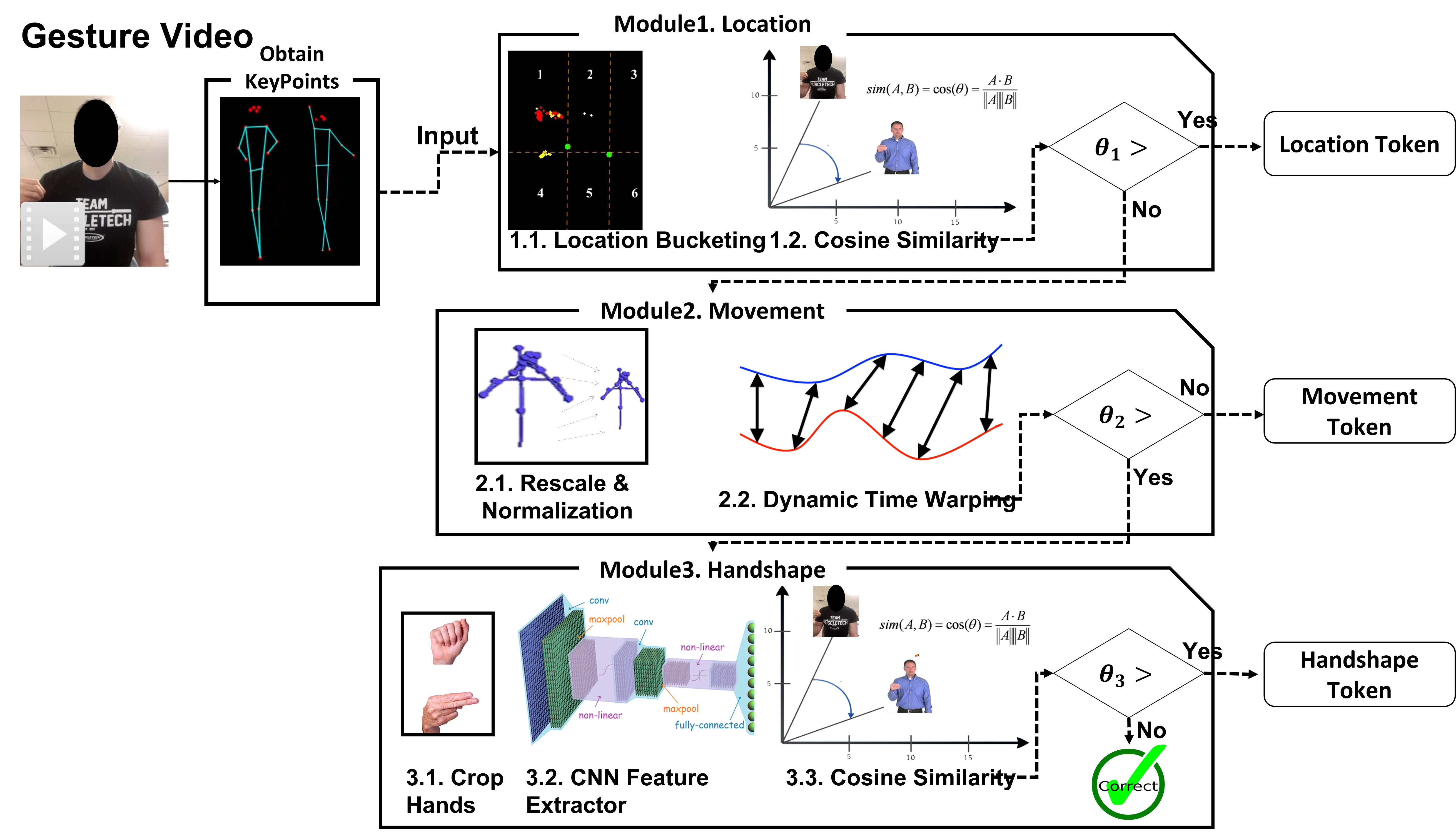}
	\caption{Token models for handshape, location and movement. }
	\label{fig:Token}
\end{figure*}
\section{Token Signature Model Development}
The models for recognizing token signatures of handshape, location, and movement are trained using examples from a well-established natural gesture-based language. 
\subsection{Handshape token model}
\label{sec:Hand}
Hand shape is a crucial element in identifying signs in ASL. To extract potential hand palm locations, we use a straightforward approach that utilizes key points obtained from pose estimation (refer to Figure \ref{fig:Token}). 
This method enables us to crop and select only the essential hand shapes at the start and end of the sign execution, while also avoiding the selection of blurry hand shapes. 
\subsection{Movement token model}
\label{sec:Move}
To train the model for movement token recognition, we utilized the manual movement segments and processed them using Posenet (as shown in Figure \ref{fig:Token}). 
The Posenet architecture generated key points for each frame, which were then temporally aligned and concatenated to form a time series. We extracted the wrist keypoint data for both hands, and calculated the $X,Y$ co-ordinate time series for each. Using these time series data, we performed second-order differentiation to obtain the wrist acceleration, velocity, and overall position. These time series were used as signatures for the movement token 


\subsection{Location token model}
\label{sec:Loc}
To learn the location tokens from the datasets, we utilized two palm locations: the start location and the end location. To identify these locations, we utilized a real-time human pose estimation model, PoseNet. This model enabled us to detect joint locations, such as the wrists, nose, eyes, elbows, and shoulders, from every frame of each video. These joint locations serve as key points for our analysis.

\section{Transfer: Sensing Technology Adaptation}
\label{transfer}
The Posenet keypoint tracking technique presented in Section \ref{sec:Move} is employed to extract the token set for wrist movement from video data. This results in a $4 \times N$ Posenet feature matrix for each gesture example, where N denotes the number of frames in the gesture. For our analysis, we consider 44 common gesture classes between the ASLText and SignFi datasets, with six replications each. A labeled set $(X^S,Y^S)$ is generated by matching the Posenet features of each gesture instance with their respective labels.

\subsection{Extracting spatial co-ordinate tracking of human wrist from WiFi}

Human hand motion in the X-Z spatial co-ordinate can be extracted from WiFi CSI if multiple antennas are available\cite{WiDraw}.  If the user's hand blocks the signal arriving at an antenna along a specific angle of attack (AoA), then the signal strength drops significantly.  Hence, if a user's hand is $r$ distance away from a receiver and causes a drop in signal strength at azimuth angle $\alpha_z$ and elevation $el$, then in the X and Z plane, the wrist co-ordinates can be computed using Equation \ref{eqn:WiDraw}.

\begin{equation}
\label{eqn:WiDraw}
x = \frac{r}{cot(\alpha_z)}, \textnormal{ and } z = \frac{r tan(el)}{cos(\alpha_z)}
\end{equation}
In the SignFi dataset, 3 transmitting and 30 receiving antennas were used to capture the human arm motion. The average $x$ and $z$ from all pairs of transmitter and receiver was computed. Each gesture instance has 200 samples of elevation, azimuth angle and distance of the hand from the sensing device. The $x,z,r$ data is computed for each such instance resulting in a $3 \times 200$ feature matrix $X^T$. This feature matrix is computed for all instances of the 44 gestures that are common across the two datasets. 

Out of the 44 gesture classes, we selected 29 gesture classes and marked then unseen. This means that during the training process of TRANSFER, we are not going to utilize the gesture labels of these 29 classes. The feature matrix from these 29 gestures form the unlabeled domain $D^u$. The feature matrix from the rest of the 15 gesture classes $\mathcal{G}_L$ form the labeled domain $D^L$.   

\subsection{Video - WiFi Latent Space Mapping}
\label{sec:ConvAuto}
\begin{figure}
	\centering
	\includegraphics[width=\linewidth,trim=0 0 0 0]{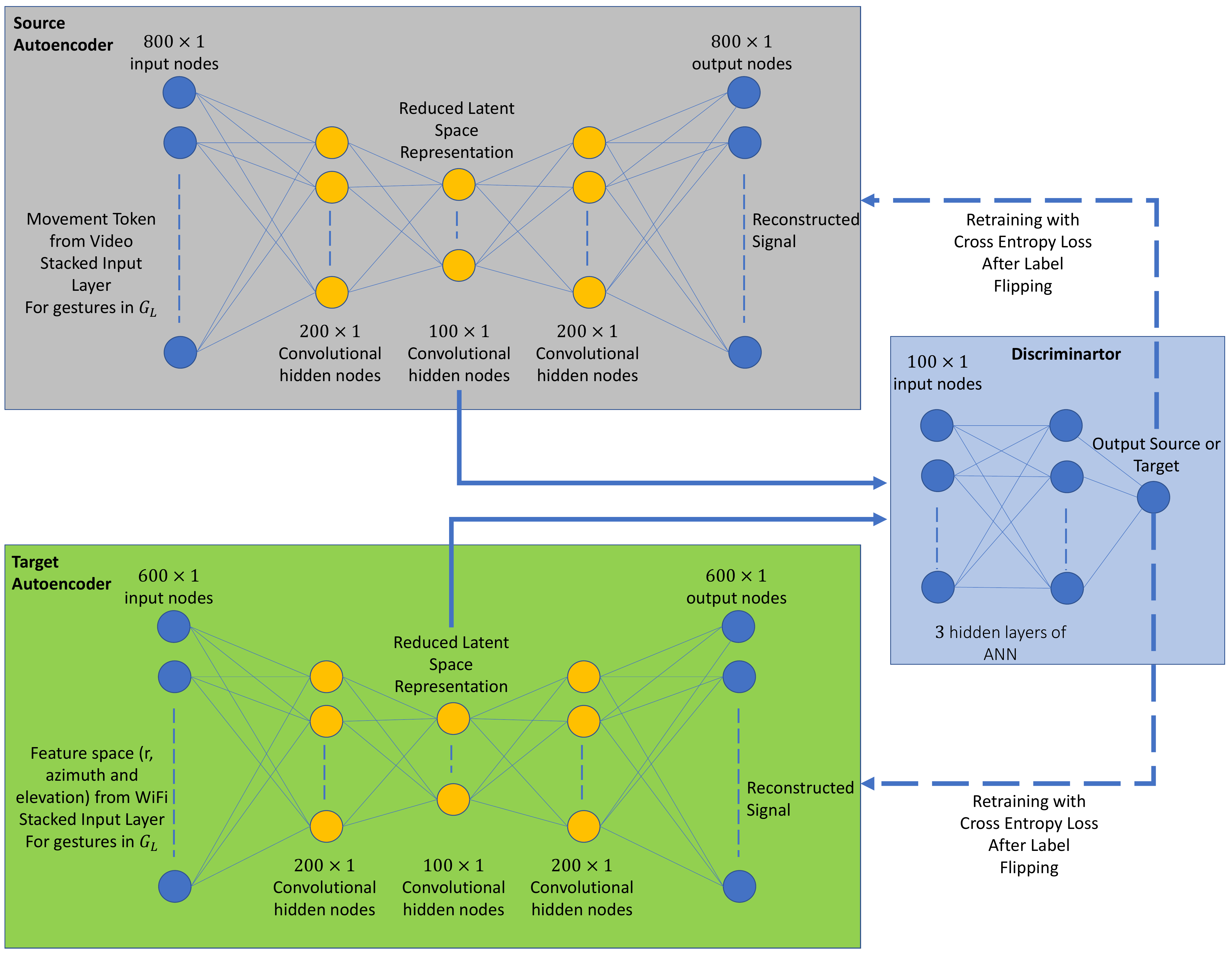}
	\caption{Latent space mapping from Video and WiFi, discriminator training to obtain technology independent latent space.}
	\label{fig:Tech}
\end{figure}
We consider a 1D CNN based autoencoder structure to extract the latent space from both video and WiFi (Figure \ref{fig:Tech}). For video, the input space is stacked across feature dimension to make it a 
4 * N input vector. For example, if we have 200 frames in a gesture then it becomes 800 * 1 input vector.
Hence the input layer has $800$ perceptrons with ReLU activation. The first hidden layer consists of a convolutional layer with 200 neurons and a stride length 2. The latent space is the second hidden layer of 100 convolutional nodes. This comprises the encoder. The decoder is the mirror image of the encoder. Similar structure is used for the WiFi data, except that the input space is stacked to a $600 \times 1$ vector.

\subsection{Discriminator training}
The discriminator is a 3 layer artificial neural network with 100 neurons in each layer with the ReLU activation function. The final output layer consists of a single neuron with $tanh$ activation function to output 1 for source technology and 0 for target technology. The training of the architecture in Figure \ref{fig:Tech} is performed in batches.

\noindent{\bf Step 1: Video autoencoder training -}To train the autoencoder for video data, we use two replications (from two different users) of each of the 15 labeled gestures in $G_L$. The training is performed for 1000 epochs using cross entropy loss.   

\noindent{\bf Step 2: WiFi autoencoder training -} We use two instances of each of the 15 labeled gestures in $G_L$ to train the autoencoder for WiFi data. The model is trained for 1000 epochs using cross-entropy loss.

\noindent{\bf Step 3: Discriminator training -} The latent space output of the video encoder is labeled as 1 while that of the WiFi encoder as 0. This training data for the initial training set from Video and WiFi is used to train the discriminator. The discriminator loss is modeled using the mean square error metric, which evaluates how accurately the discriminator can classify 1 / 0. The initial run took 330 epochs to reach a low loss threshold of 0.01. 

\noindent{\bf Step 4: Label flipping and loss computation - } The labels of the video latent features are computed for the fourth user and the label is flipped to 0. Similarly the WiFi latent features are computed for the third user and the label is flipped to 1. The label flipping is applied such that, for example, when the discriminator will  classify the video latent features with label 1, the backpropagation process used to update the model weights will see this as a large error and will update the model weights to correct for this error, in turn making the source video autoencoder better at generating technology invariant features.

\noindent{\bf Step 5: TA model - } The TA model takes three models, source autoencoder, target autoencoder and discriminator, as input. This function combines these three models to create a new TA model. The weights in the discriminator are set as not trainable, which only affects the weights seen by the source and target autoencoder models, but not the standalone discriminator model.
The training process stops when the discriminator loss reaches a high threshold of 0.5. This took 7 iterations of Step 3 through 5 for the Video-WiFi example. Since we had only 6 users in ASLTEXT and 5 users in SignFi, we chose to repeat users starting from user 1. But we used different executions for each user. 

\subsection{Test Gesture recognition using transductive fine tuning}

To recognize gestures from the unlabeled domain $\mathcal{D}^u$, we utilize transductive fine tuning, as described in~\cite{OneFi}. This involves fixing the weights of the encoder for both the video and WiFi data. We first pass a test gesture instance $x^u_i \in X^u$ through the WiFi encoder to obtain its latent space representation, denoted by $\phi^T(x^u_i)$. Then, for each of the 29 unseen gesture classes, we pass the corresponding video samples through the video encoder to generate the video latent space representation $\phi^S(X^S_{\mathcal{G}/\mathcal{G}L})$. We evaluate the cosine similarity between $\phi^T(x^u_i)$ and each element of $\phi^S(X^S{\mathcal{G}/\mathcal{G}L})$. The gesture class corresponding to the element in $\phi^S(X^S{\mathcal{G}/\mathcal{G}_L})$ with the highest cosine similarity is recognized as the class of the test gesture instance.

\section{Evaluation and Results}

In this section, we showcase the effectiveness of \textit{TRANSFER} in enhancing the cross-domain recognition of SignFi gestures by extracting technology-invariant features from the ASLTEXT video dataset. To evaluate this task, we employ two additional approaches as comparators: a) a replication of the few-shot gesture recognition approach OneFi~\cite{OneFi}, and b) another implementation of \textit{TRANSFER} without the language-driven dimensionality reduction method. 
\subsection{Datasets}
\label{sec:data}
The following datasets were used in this project:

\noindent{\bf SignFi dataset:} SignFi has CSI data for 276 ASL gestures. Each gesture was performed by five users. 20 repetitions of each gesture is available. CSI data between three transmitting and 30 receiving antennas are available for each gesture execution. A total of 5520 instances of 4D CSI data are extracted from this dataset.

\begin{figure*}[t]
	\centering
	\includegraphics[width=\linewidth,clip=true, trim=0 0 0 0]{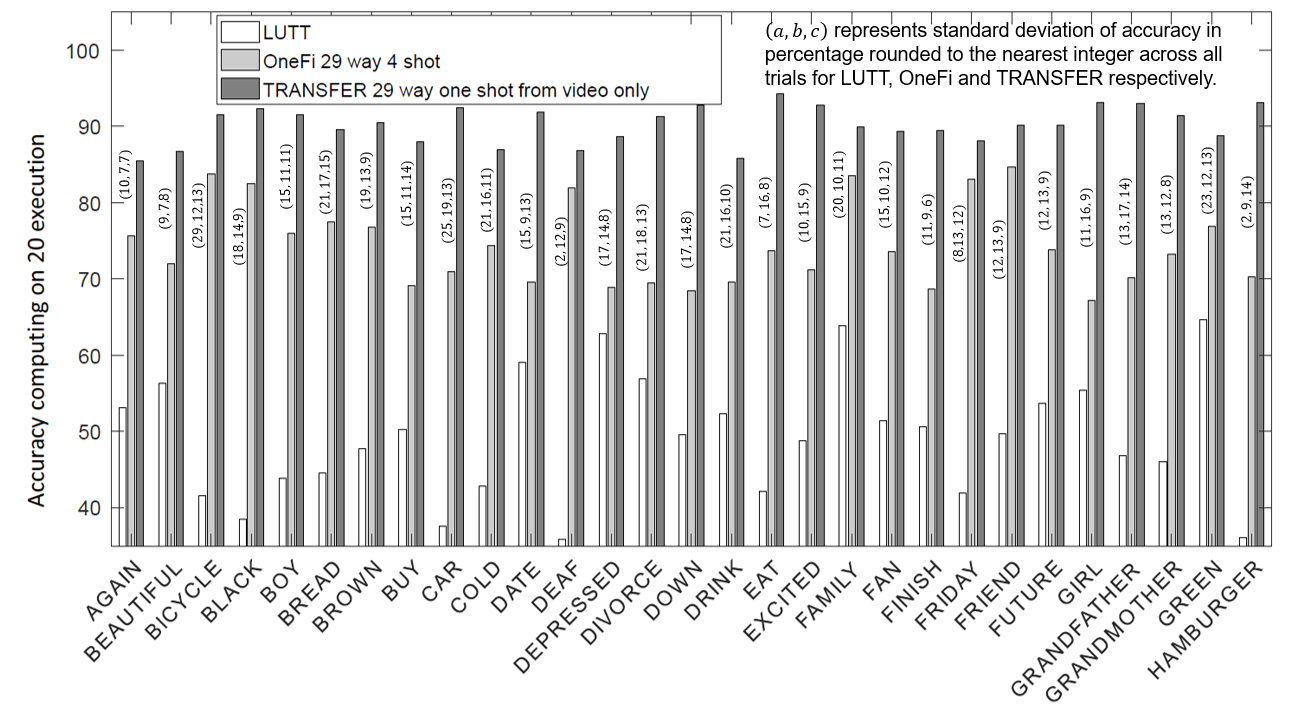}
	\caption{Recognition results for Signfi WiFi dataset }
	\label{fig:SignFi}
\end{figure*}

\noindent{\bf ASLText dataset:} The ASLText dataset~\cite{bilge2019zero} is derived from the ASLLVD dataset collected at the Boston University. It consists of video data (33 frames per second) of 250 ASL words signed by six native signers with varied background. It consists of 1598 videos where each video is approximately a second long.   



\noindent{\bf SCEPTRE dataset:} This dataset consists of accelerometer, gyroscope, orientation and electromyogram (EMG) sensor data for 23 ASL gestures published by Paudyal et al~\cite{Paudyal16Sceptre}. It has 20 users and the data was collected using Myo device. The average length of each example gestures is 3s. The sampling rate of the accelerometer data was 20 Hz, while that of EMG data is 50 Hz. 

\noindent{\bf IMPACT dataset:} This dataset consists of video data from 10 users on 23 ASL gestures similar to SCEPTRE published by Kamzin et al~\cite{kamzin2021concept}. The average length of each example gesture is also 3s. The video data is collected using smartphones at 33 fps. 

\subsection{Replication of OneFi}
\noindent{\bf Justification:} By comparing our results to OneFi~\cite{OneFi}, we can demonstrate the effectiveness of using video sensing technology to enhance cross-domain recognition accuracy. OneFi is a few-shot recognition mechanism that employs a combination of multi-head attention-based transformer models proposed in~\cite{vaswani2017attention} and transductive fine-tuning to recognize hand gestures using WiFi. It reports an overall accuracy of 84.2\% on one-shot recognition of 40 hand gestures from 10 users, with only six gestures being unseen. However, the reported unseen gestures are relatively simple, involving movements that can be composed using multiple linear submovements. In contrast, the gestures in ASLTEXT may involve complex motions, such as arcs, shakes, and circular motions with directional discontinuities. Therefore, our approach represents a significant advance in cross-domain gesture recognition.

\noindent{\bf Replication:}  The Doppler spectrogram features used in OneFi~\cite{OneFi} are computed from the raw 4D CSI data in the SignFi~\cite{ma2018signfi} dataset. We use all the 30 subcarriers available in the SignFi data. The DC offset of each subcarrier was removed. Phase offset was computed by taking two antennas at random and computing the conjugate multiplication of the CSI reading. Each processed CSI is then passed through the high pass filter and short-term fast fourier transform (sFFT) was computed on each CSI data. We then computed the velocity doppler mapping using Equations 8,9, and 10 in ~\cite{OneFi}. This acts as an input to the WiFi Transformer. The WiFi transformer was implemented using the open source code provided in \url{https://github.com/tensorflow/tensor2tensor/}. We used the \texttt{mtf\_transformer2.py} since the velocity doppler mapping is a 1D timeseries. For the transuctive fine tuning component, we used the same code as developed for \textit{TRANSFER}. 

We did not replicate the virtual gesture generation process described in OneFi. Instead, to achieve a fair comparison, we took four executions of each of the 44 gestures used for \textit{TRANSFER} and trained the WiFi transformer model. Hence, we compare the 29 way 4 shot configuration of OneFi~\cite{OneFi} with 29 way one shot configuration of \textit{TRANSFER}. We call this one shot although we do not use any WiFi data from SignFi for these 29 gestures to train \textit{TRANSFER}. We only use video data of these 29 gestures in the testing phase.
\subsection{Language Unaware Technology Transfer}
\label{sec:LUTT}
The Language Unaware Technology Transfer (LUTT) method does not utilize any language-based techniques to reduce dimensionality. Instead, it employs a video autoencoder, which has a 3D convolution layer as the initial hidden layer, in contrast to a 1D convolutional neural network (CNN). The input layer is also 3D in shape, with dimensions of $256 \times 256 \times 200$, and accepts videos as input. After the first 3D convolutional layer, a second 3D convolutional layer with dimensions of $64 \times 64 \times 20$ follows. This layer is then flattened into a 2D fully connected layer of size $150 \times 60$, which is accomplished by including four rectified linear units (ReLU) between the 3D and 2D layers. The 2D layer is further flattened into a 1D fully connected layer of size $100 \times 1$, which constitutes the latent space. The decoder architecture is a mirror image of the encoder architecture. The training process for LUTT is identical to that of the \textit{TRANSFER} method.

\begin{table}[t]
	\sffamily
	\centering
	\footnotesize
	\caption{Summary of \textit{TRANSFER} Accuracy for tasks }
	\scriptsize
	\begin{tabular}{p{1.7 in}|p{0.5 in}|p{0.8 in}} \toprule
\textbf{Task}		&{\bf Accuracy} & {\bf Improvement}\\ \midrule
		Recognition of 29 common gestures between ASLText and SignFi (Training using ASLText and Recognition on SignFi) & 90.87\% & 16\% better than a replication of OneFi~\cite{OneFi}\\
		Recognition of 7 common gestures by transferring knowledge from Video to accelerometer sensing domain & 82.1\% & first attempt \\
		 Recognition of 7 common gestures by transferring knowledge from accelerometer to WiFi sensing domain & 79\% & second attempt \\
		\bottomrule
	\end{tabular}
	\label{tbl:Sum}
\end{table}
\subsection{Knowledge transfer across technology}
\noindent{\bf WiFi gesture recognition with video based learning:} As discussed  in Section \ref{transfer}, the SignFi and the ASLText only have the movement and location tokens common in the alphabet.  Therefore, we have exclusively utilized the movement and location tokenizer, spatial co-ordinate interpreter, and token recognition modules of \textit{TRANSFER} framework. Figure \ref{fig:SignFi} depicts the outcomes obtained by applying the \textit{TRANSFER} method to the SignFi dataset. We have solely employed the ASLText dataset for training purposes. The recognition accuracy for the 29 gestures within the unlabelled SignFi dataset is presented. Each gesture within the SignFi dataset has 20 repetitions across 5 users. The average accuracy for the 29 unlabeled gestures is 90.87\%. Additionally, Figure \ref{fig:SignFi} displays the accuracy for individual gestures, alongside the accuracy of the other two comparators. Notably, the accuracy of OneFi has decreased dramatically on all 29 gestures, dropping from 84.2\% (as reported in the paper~\cite{OneFi}) to 74.87\% in the SignFi dataset. This may be attributed to two factors: a) poor cross-domain generalization by the WiFi transformer, or b) the SignFi gestures are more complex, hence the velocity doppler map represents an inadequate feature space. This highlights that the use of knowledge from videos enhances the accuracy by almost 16\%. The accuracy of LUTT is also illustrated in Figure \ref{fig:SignFi}. However, it is significantly lower than that of \textit{TRANSFER}. The principal reason for this is that the 3D CNN-based autoencoder is overfitting to the available data. This is evident from the low training loss but the very high validation loss. This demonstrates the benefits of using language-driven dimensionality reduction techniques. 

The most time-consuming process in \textit{TRANSFER} is the training phase. For the Video to WiFi task, the discriminator training application utilized an Intel Core i7 10th Gen system with 8GB RAM and 1 TB hard disk. No GPU was employed for this purpose. The training comprised 1200 epochs, with each epoch involving the backpropagation training of the two autoencoders, label flipping, autoencoder weight freeze, and backpropagation on the discriminator. The entire training process for the 15 labeled gestures took 26 hours and 34 minutes. However, this time can be substantially reduced by utilizing a GPU, which was not investigated in this study. On the same machine configuration, the testing time is considerably lower, with an average of 178 seconds for each unseen gesture.

\section{Generalizability of \textit{TRANSFER}}

To demonstrate the versatility of \textit{TRANSFER}, we present two additional examples: a) video as the source domain and accelerometer as the target domain (again due to limitation of accelerometer sensing we are only confined to transferring knowledge about movements) and b) accelerometer as the source domain and WiFi as the target domain. 

\subsection{\textit{TRANSFER} from Video to Accelerometer}
\label{sec:AccelV}
The same architecture as shown in Fig. \ref{fig:Tech} is used for knowledge transfer between video and accelerometer. The autoencoder structure for video remains the same. Further, the autoencoder architecture for WiFi can be reused for accelerometer. Instead of using the WiFi features, we use the 3D displacement vector extracted from the accelerometer as described in Marzia et al~\cite{cescon2021activity}. The accelerometer is sampled at 20 Hz and a 3 second gesture duration is considered. The accelerometer data is upsampled to a length of 600 samples. The upsampling is performed using polynomial spline interpolation. The gesture data of 600 samples is then used as input to the accelerometer autoencoder, which has the same structure as shown in the Target Autoencoder block in Fig. \ref{fig:Tech}. The structure of the discriminator also remains the same.

Results:  23 ASL signs are common between the SCEPTRE (accelerometer)~\cite{Paudyal16Sceptre} and the IMPACT (video) datasets~\cite{kamzin2021concept}. We utilize 16 gestures to train the TRANSFER architecture in Fig. \ref{fig:Tech} and form the labeled domain. The other 7 gestures are marked as unseen and form the unlabeled domain. Each gesture has 10 replications in video and 20 replications in accelerometer. For the 16 gestures in the labeled dataset, we use all user data for training the video and accelerometer autoencoder. The \textit{TRANSFER} framework is then tested for recognition of the other 7 gestures from the 20 users in the SCEPTRE dataset. Table \ref{tbl:accel} shows that the average recognition accuracy on the 7 gestures is 82.1\%. This reduction in accuracy can be attributed to the much lower amount of training data available as compared to ASLText and SignFi. 
 
\begin{table}[t]
	\sffamily
	\centering
	\footnotesize
	\caption{Gesture recognition results for transfer of knowledge from Video to accelerometer }
	\scriptsize
	\begin{tabular}{p{1.7 in}|p{0.5 in}|p{0.5 in}|p{0.5 in}} \toprule
\textbf{Gesture}		&{\bf Number of instances} & {\bf Number Recognized} & {\bf Accuracy}\\ \midrule
		about & 20 & 17 & 85\%\\
		cop & 20 & 19 & 95\% \\
		decide & 20 & 14 & 70\% \\
		father & 20 & 15 & 75\%\\
		goout & 20 & 17 & 85\%\\
		hospital & 20 & 18 & 90\%\\
		tiger & 20 & 15 & 75\%\\
		\bottomrule
	\end{tabular}
	\label{tbl:accel}
\end{table}

\subsection{\textit{TRANSFER} from accelerometer to WiFi}
In this example, we use the common gestures between SignFi and SCEPTRE to show knowledge transfer from accelerometer to WiFi. For this purpose, we utilize the TRANSFER framework shown in Fig. \ref{fig:Tech}, and replace the video autoencoder by the accelerometer autoencoder used in Section \ref{sec:AccelV}. There are 18 common gestures between the SCEPTRE and SignFi dataset. We use 13 common gestures from both the datasets to train the modified \textit{TRANSFER} framework. Table \ref{tbl:WiFiAccel} shows the accuracy of recognition for the rest of the 5 gestures. As seen in the Table \ref{tbl:WiFiAccel} the accuracy is 79\% and further drops from the Video to WiFi or Video to accel transfer objectives. There may be two contributing factors: a) lack of training data, and b) the accelerometer as a source signal may not have rich features to exploit for the knowledge transfer mechanism.

\begin{table}[t]
	\sffamily
	\centering
	\footnotesize
	\caption{Gesture recognition results for transfer of knowledge from accelerometer to WiFi }
	\scriptsize
	\begin{tabular}{p{1.7 in}|p{0.5 in}|p{0.5 in}|p{0.5 in}} \toprule
\textbf{Gesture}		&{\bf Number of instances} & {\bf Number Recognized} & {\bf Accuracy}\\ \midrule
		help & 100 & 83 & 83\%\\
		sorry & 100 & 77 & 77\% \\
		cat & 100 & 71 & 71\% \\
		good & 100 & 85 & 85\%\\
		happy & 100 & 82 & 82\%\\
		\bottomrule
	\end{tabular}
	\label{tbl:WiFiAccel}
\end{table}


\section{Limitations of \textit{TRANSFER}}
There are several limitations of the \textit{TRANSFER} approach as detailed below: 
 
\noindent{\bf Movement only:} In this paper, we could show knowledge transfer across sensing technology by only considering movement. This is because the non-video sensing techniques such as WiFi and accelerometers do not capture the handshape and finger movements. They only capture the wrist movements. 
The transfer of knowledge about the handshape across sensing domain is still an issue under investigation.

\noindent{\bf Initialization overhead:} One of the main drawbacks of \textit{TRANSFER} is that it requires common grammar. However, given that the two sensing technology are collecting data for the same application domain, there has to be a sensor independent common concept. The task of the user is to recognize such concepts and use that in \textit{TRANSFER}. For the ASL recognition application, the concepts are obtained through consultation of ASL linguistics experts. Such concepts in an application has to be determined through expert collaboration and is an overhead for this technique.  

\noindent{\bf Non-gestural domain adaptation:} In this paper, we have only shown the execution of \textit{TRANSFER} on gesture recognition application. However, the \textit{TRASFER} technique is intended to be applied in non-gestural application domain as well. This is an item of future research.

\section{Conclusions}

In this paper, we demonstrate \textit{TRANSFER} framework for identification and recognition of gestures by transferring the knowledge across technology and application domain. The advantage of \textit{TRANSFER} is that lack of training data for a particular technology can be overcome by utilizing knowledge from widely available video data. Moreover, unseen gestures with definitions can be recognized with an already pre-trained machine on ASL Vocabulary of gestures in terms of handshape, location and movement. Application of \textit{TRANSFER} on WiFi based device free gesture recognition shows 91\% accurate recognition without using any part of the WiFi dataset in training phase. This is a ground breaking innovation that can make WiFi based device free gesture recognition widely applicable across many gestures and users. 

Although we have focused on the vast domain of gesture recognition, it can be observed that the core methodology of \textit{TRANSFER} does not assume any particular source or target technology. Using \textit{TRANSFER}, we can envision knowledge transfer between - video and accelerometry for action recognition, and echocardiography (EchoG) images and eletrocardiogram (ECG) signals to improve the accuracy of detection of cardiac diseases using low cost ECG.  

\bibliographystyle{ACM-Reference-Format}
{00}
\end{document}